\documentclass{article}

\usepackage[dblblindworkshop, final]{neurips_2025}

\usepackage[utf8]{inputenc} 
\usepackage[T1]{fontenc}    
\usepackage{hyperref}       
\usepackage{url}            
\usepackage{booktabs}       
\usepackage{amsfonts}       
\usepackage{nicefrac}       
\usepackage{microtype}      
\usepackage{xcolor}         
\usepackage{amsmath}
\usepackage{graphicx}
\usepackage{subcaption}
\usepackage{float}        
\usepackage[section]{placeins}
\usepackage{algorithm}
\usepackage{algpseudocode}
\usepackage{bbm}

\title{Delay-of-Gratification as a Multi-Agent Survival Micro-benchmark for Long-Horizon LLMs: Social Exposure, Personas, and Tool Use Budgets}

\author{%
  Olga Manakina \\
  Department of Cognitive Science\\
  Carleton University, 
  Ottawa, ON K1S 5B6 \\
  \texttt{olgamanakina@cmail.carleton.ca} \\
  \And
   Igor Bogdanov \\
   Systems \& Computer Engineering \\
   Carleton University,
   Ottawa, ON K1S 5B6 \\
   \texttt{igorbogdanov@cmail.carleton.ca} \\
  \And
  Chung-Horng Lung \\
  Systems \& Computer Engineering \\
  Carleton University,
  Ottawa, ON K1S 5B6 \\
  \texttt{chlung@sce.carleton.ca} \\
}
 
\begin{document}

\maketitle

\begin{abstract}
Large language models (LLMs) are increasingly deployed as multi-turn agents that must sustain goals, use tools, and adapt to other agents over extended interactions. However, existing research lacks auditable, multi-turn, multi-factorial experiments that quantify LLM behavior under explicit constraints, with time-resolved statistics that reveal how behavior unfolds over long horizons. To address this gap, we develop a multi-agent Micro-benchmark inspired by the Stanford marshmallow experiment: ReAct agents operate minute-by-minute with a "raise a question" tool under a per-step budget, while we factorially manipulate social context (broadcast vs.\ isolated), personas (age, hedonic drive), and metacognitive policy (must vs.\ may follow instructions). We analyze outcomes with Kaplan--Meier (KM) survival curves and discrete-time hazard models over a long risk horizon. Across 19{,}200 agent trajectories in 64 cells (horizon $T=19$), $99.9\%$ of runs were valid. Behavior shows a sharp early "eat" impulse (initial eat $=0.125$), a total eat rate $=0.241$, and $75.9\%$ of agents persist to the end; the waiting profile is summarized by median time-to-eat $\approx 14.8$ and RMST $\approx 14.8$. In a discrete-time hazard model, isolation reduces per-minute risk relative to broadcast (OR $=0.78$, $95\%$ CI [0.73, 0.83], $p<.001$), whereas a \textsc{must}-use self-questioning policy increases risk (OR $=1.42$, [1.35, 1.50], $p<.001$). Hedonic and age personas strongly modulate risk: vs.\ \textit{crave}, \textit{like} (OR $=0.28$), \textit{none} ($0.19$), and \textit{neutral} ($0.03$) reduce hazard; vs.\ \textit{adult}, \textit{child} increases hazard (OR $=66.3$) and \textit{senior} is elevated ($7.55$) (all $p<.001$). On average, agents ask $\approx 7.12$ questions and hit the per-step budget in $\approx 6\%$ of minutes; question-asking declines faster under broadcast than isolation. Further ablation experiments demonstrated that removing hedonic drive and/or persona age systematically increases survival and completion, narrows the broadcast/isolated gap, and leaves the \textsc{must} vs.\ \textsc{may} ordering intact (\textsc{must} is riskier); the combined ablation (no hedonic $+$ no persona age) yields the highest completion (approaching $1.0$) and distinct tool-usage dynamics with higher initial questioning rates that gradually decrease over time. These results establish delay-of-gratification as a compact multi-turn interaction benchmark that captures social contagion and tool-use dynamics in LLM agents, offering a reproducible testbed and statistics to analyze long-horizon, multi-agent behavior. 

\end{abstract}
\section{Introduction}
Modern uses of large language models (LLMs) are inherently conversational and iterative, as users and agents co-construct tasks over multiple turns, revise goals, and recover from mistakes. Recent multi-turn evaluations show that single-turn prowess does not guarantee long-horizon reliability: agentic setups reveal gaps in reasoning and decision-making \cite{liu2023agentbench}, tool use and natural-language feedback help but interact idiosyncratically with training and instruction tuning \cite{wang2023mint}, and performance can drop substantially when moving from single- to multi-turn interaction \cite{laban2025lost}. These observations motivate \emph{auditable, constrained, multi-factorial} experiments that measure how agent behavior unfolds over time and in the presence of other agents.\\

\textbf{Hypotheses: }Inspired by a classic Stanford study on delayed gratification \citep{mischel1972cognitive} and by recent LLM studies replicating marshmallow-like scenarios \citep{coletta2024subrational} and other cognitive tasks~\citep{lampinen2024content, strachan2024tom}, we test five hypotheses in a controlled Micro-benchmark with a minimal action space. 
\textbf{H1} \emph{Social visibility}: when agents can observe peers, the hazard of committing to the immediate option increases relative to isolation.
\textbf{H2} \emph{Internal state}: personas reflecting stronger hedonic drive and child age elevate hazard, whereas neutral drive and adult age reduce it.
\textbf{H3} \emph{Metacognition}: a mandatory self-questioning step (\textsc{must}) changes hazard relative to optional use (\textsc{may}); we assess whether such scaffolding stabilizes behavior.
\textbf{H4} \emph{Temporal structure}: the per-minute hazard is \emph{non-constant} across the horizon (i.e., behavior displays systematic time dependence), without pre-specifying its shape.
\textbf{H5} \emph{Prompt crafting} \emph{pre-registered expectation}: more prescriptive prompt scaffolding and instruction complexity, including enforced metacognitive steps, should improve adherence (lower hazard) compared to a minimalist design.\\

\textbf{Brief results:}
Across a 19-minute horizon with 19{,}200 trajectories, we find strong time dependence (\textbf{H4}) and that social visibility raises risk (\textbf{H1}); isolation lowers per-minute hazard vs.\ broadcast (OR $\approx 0.78$). Personas strongly stratify outcomes (\textbf{H2}); a \textsc{must} policy \emph{increases} hazard (OR $\approx 1.42$), indicating metacognitive enforcement can backfire (\textbf{H3}). Contrary to \textbf{H5}, heavier prompt scaffolding does not uniformly help and can degrade reliability in this setting. Our ablation experiments show three key effects of removing persona components (hedonic drive and/or age): (1) systematically improved survival and completion rates, (2) reduced differences between broadcast and isolated conditions, while preserving the higher risk of mandatory tool use, and (3) in the case of complete ablation (no hedonic or age), near-perfect completion rates and distinctive tool-usage patterns marked by increased early questioning that gradually diminishes over time. These results clarify where multi-turn agents fail and how social context and scaffolding shape behavior over time. 

\section{Related work}

\textbf{Classic Cognitive Tasks, LLMs and Multi-Turn Interactions.} Our study contributes to a recent body of work that adapts classic cognitive tasks to investigate LLM capabilities. Models show human‑like \emph{content effects} in reasoning \citep{lampinen2024content}; near‑human performance on Theory‑of‑Mind tasks can degrade under prompt variations \citep{strachan2024tom,kosinski2024tom}; and judgment, decision‑making, and memory studies report framing/probability biases and capacity limits \citep{binz2023cognitive,wang2024linda,zhang2024wmreasoning,gong2024wmcapacity}. These studies are largely single-prompt or short-horizon; we instead target \emph{multi-turn} interaction by importing delayed gratification into a controlled, minute-by-minute, multi-agent setting.

\textbf{Human delay of gratification and intertemporal choice.}  The Stanford marshmallow experiments and subsequent studies on delay of gratification show that attention and cognitive strategies modulate waiting, inspiring the "hot/cool" model of self‑control \citep{mischel1972cognitive,metcalfe1999hot}. Long‑term links to outcomes are moderated by environmental reliability and socioeconomic context \citep{kidd2013rational,watts2018revisiting}, with neural work implicating adult self‑control circuitry \citep{casey2011behavioral}.  A recent study explored marshmallow‑like scenarios in the context of LLMs \citep{coletta2024subrational}, although it did not include temporal/social analyses. In behavioral economics, intertemporal choice formalizes conflicts between immediate and delayed rewards via hyperbolic discounting and time-inconsistent preferences (where immediate rewards are disproportionately valued over future ones, formalized in the $\beta$-$\delta$ model), commitment, and naïve vs.\ sophisticated agents \citep{ainslie1992picoeconomics,laibson1997golden,odonoghue1999doing}; classic procedures quantify delay preferences \citep{mazur1987adjusting}. 
We adapt these insights to an LLM survival‑analysis frame over discrete minutes.

\textbf{Scaffolding, multi-agent interaction, evaluation, and personas.} Our framework builds on several key developments in LLM interaction design and evaluation. Reasoning architectures such as ReAct, Self-Ask, and Reflexion scaffold stepwise deliberation and tool use \citep{yao2023react,press2022measuring,yao2023tree,shinn2023reflexion}. Multi-agent coordination and social simulation, e.g., debate, role-based systems, and long-horizon agent societies---provide structure for interaction and influence \citep{du2023improving,li2023camel,wu2024autogen,park2023generative}. Our analysis uses time-to-event tools, KM and discrete-time logistic hazard models, to quantify factor effects on waiting \citep{kaplan1958nonparametric,singer1993makes,allison1982discrete}. Finally, persona prompting distinguishes role-play from personalization \citep{tseng2024two}; although personas may not improve objective task performance and can bias behavior \citep{zheng2024personas}, we employ them as controlled manipulations while acknowledging the limitations of LLMs as human surrogates \citep{gao2025caution}.

\subsection{Contributions}

We reframe the classic marshmallow test as a discrete-time, long-horizon survival task to evaluate decision-making in LLM agents. Our primary contribution is a controlled multi-agent experimental framework, formalized as a finite-horizon MDP (isolated) and POMDP (broadcast), paired with rigorous survival-based evaluation methods. We implement a factorial design manipulating agents' social context (isolated vs.~broadcast peer exposure), internal personas (hedonic drive, age), and metacognitive scaffolding (mandatory vs.~optional internal tool use, subject to a per-step question cap). Using Kaplan–Meier curves and discrete-time logistic hazard models with cluster-robust standard errors, we show that peer visibility significantly increases agents' risk-taking (higher hazard of early consumption). Internal persona prompts strongly influence temporal decision dynamics, with child-like and craving personas elevating early-eating hazards. Counterintuitively, a mandatory metacognitive intervention (forced tool use) increased the hazard of giving into temptation. Our findings highlight the critical role of multi-turn, socially contextualized evaluation environments for understanding intricate agent behaviors over extended decision horizons.

\section{Experiment Setting}

\begin{figure}[h!]
    \centering
    \begin{subfigure}[b]{0.49\textwidth}
        \centering
        \includegraphics[width=\textwidth]{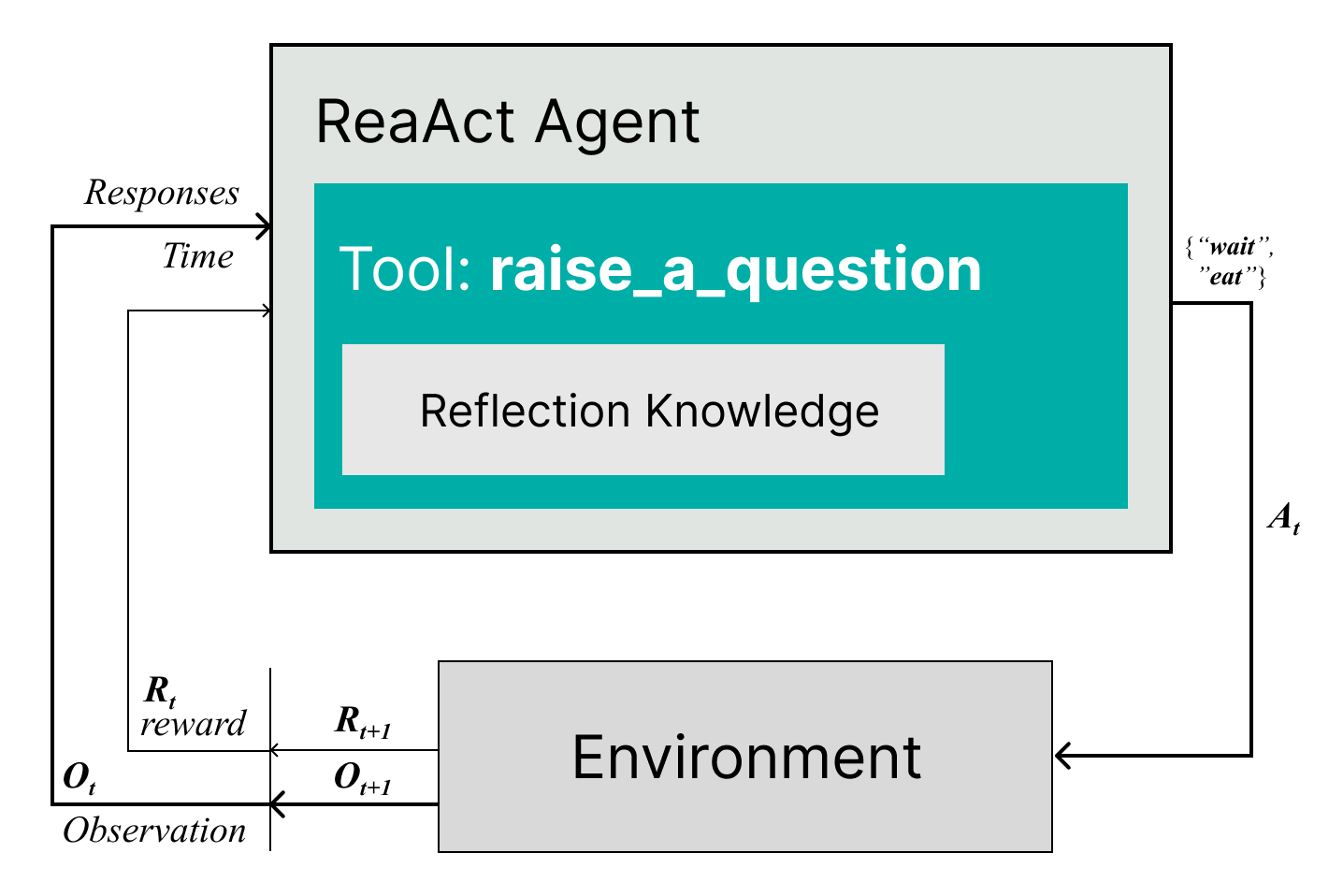}
        \caption{Episode Modeling as POMDP.}
        \label{fig:pomdp}
    \end{subfigure}
    \hfill
    \begin{subfigure}[b]{0.49\textwidth}
        \centering
        \includegraphics[width=\textwidth]{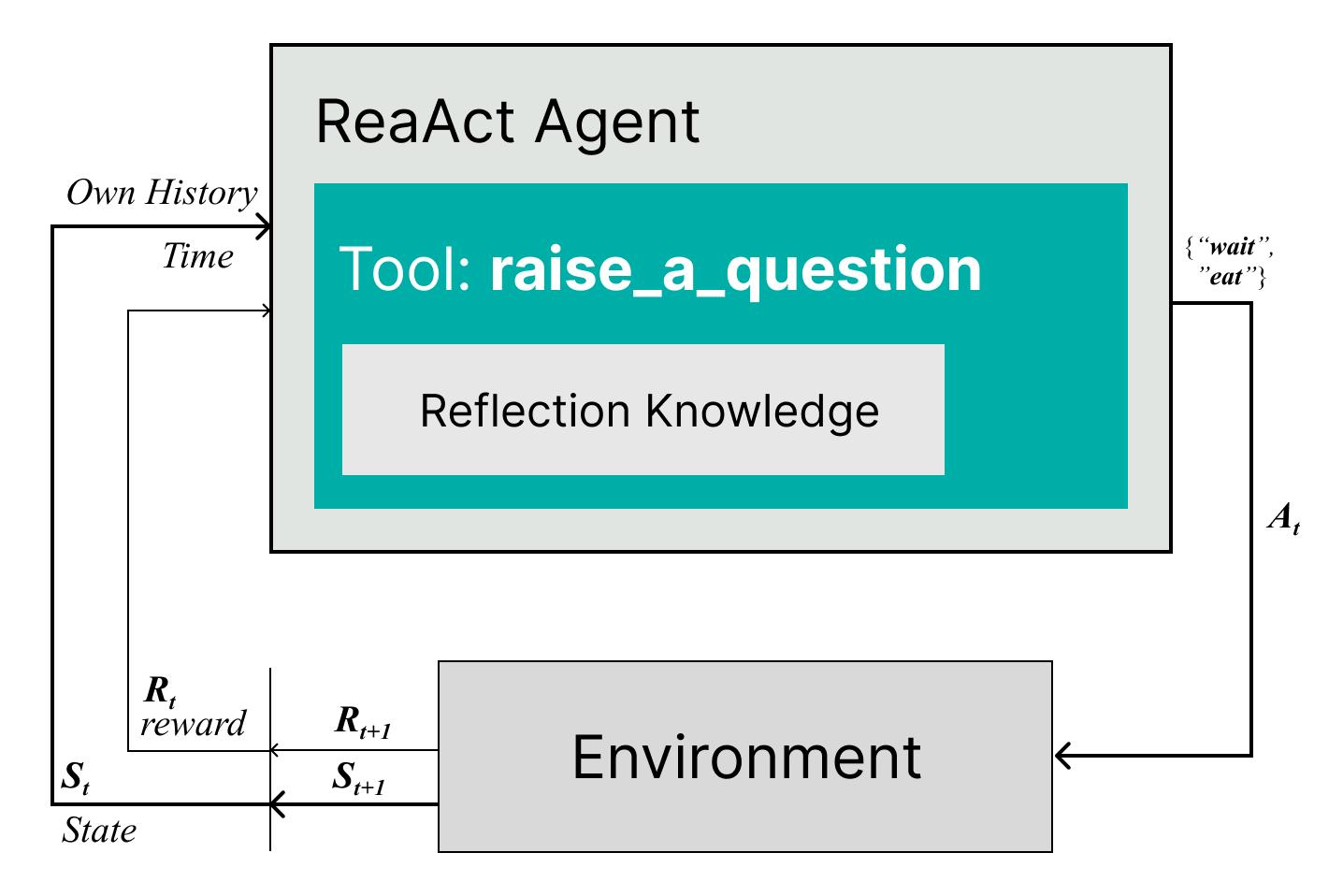}
        \caption{Episode Modeling as MDP.}
        \label{fig:mdp}
    \end{subfigure}
\caption{
Interaction loops for (a) the partially observable Markov decision process (POMDP, broadcast condition) and (b) the fully observable Markov decision process (MDP, isolated condition). 
In (a), the agent observes time and peer responses, introducing partial observability. 
In (b), the agent observes only time and its own history, rendering the environment fully observable. 
At each step, the agent internally uses a capped \texttt{raise\_a\_question} tool, 
then chooses an action $a_t \in \{\text{wait}, \text{eat}\}$. 
The environment provides a reward, updates the state, and advances to the next observation.}
    \label{fig:mdp_diagrams}
\end{figure}

\paragraph{Environment and Episode Modeling.}
We evaluate LLM agents in a finite-horizon, multi-turn environment formalized as a Partially Observable Markov Decision Process (POMDP) Figure~\ref{fig:pomdp} \citep{Kaelbling1998}, characterized by horizon \( H = \textit{risk\_horizon}+1 \). At each step \( t \in \{0,\dots,\textit{risk\_horizon}\} \), the environment state \(S_t\) evolves based on the agent's action \(A_t\). The agent receives an observation \( O_t \), composed of the current step index and, in broadcast conditions, recent peer actions (\(\texttt{others\_responses}_t\)). After optional internal deliberation via the \texttt{raise\_a\_question} tool (limited by a per-step budget), the agent emits a constrained action \( A_t \in \{\text{I wait},\text{I eat the marshmallow}\} \). The environment returns a reward \( R_{t+1} \), increments the step \( t \rightarrow t+1 \), and provides the next observation \( O_{t+1} \). Agents reaching the end of the risk horizon without eating move to the threshold step (\(\textit{threshold\_step}=\textit{risk\_horizon}+1\)), receiving the delayed payoff. Formally, the interaction loop at each step is:
\[
\small
\begin{aligned}
O_t&=[\text{Time}(t),\;\mathbbm{1}_{\text{broadcast}}\cdot\texttt{others\_responses}_t],\\[2pt]
A_t &= \pi_{\theta}\left(O_t,\;\{\texttt{raise\_a\_question}(O_t,i)\}_{i=1}^{k_t}\right),\quad 0\le k_t\le\textit{cap},\\[2pt]
R_{t+1}, S_{t+1}, O_{t+1} &= \mathcal{E}(S_t, A_t),\\[2pt]
b_t(S_t) &= P(S_t \mid O_{0:t}, A_{0:t-1}),
\end{aligned}
\]
where the tool is \textit{internal} and does not alter \( S_t \), and in isolated conditions (\(\mathbbm{1}_{\text{broadcast}}=0\)), observations \(O_t\) fully determine the underlying state \(S_t\), reducing the environment to a finite-horizon Markov Decision Process (MDP) Figure~\ref{fig:mdp} \citep{Puterman1994}.

The agents operate within a ReAct loop \citep{yao2023react} (Thought + Tool → PAUSE → Observation → Thought + Answer) with a dedicated validation tool, \texttt{raise\_a\_question}, which is gated by a per-step budget. The environment is designed to be turn-based and synchronous: at each minute, all active agents observe, decide, and act. When an agent chooses to eat, they are eliminated from subsequent minutes, while waiting maintains the agent's participation but keeps them at risk.

We implement a factorial design that manipulates agents' social context (isolated vs. broadcast), internal personas (hedonic drive and age), and metacognitive scaffolding (mandatory vs. optional tool use). We assess these factors via KM curves \citep{kaplan1958nonparametric} and discrete-time hazard models \citep{allison1982discrete}.

\textbf{Time Horizon \& Rewards: }
The scenario maps 20 minutes of delay gratification to $T$ discrete steps (default $T = 20$, with Step 0 serving as initialization). As LLMs do not have an inherent concept of time, the time is modeled in natural language: at each step, the agent is reminded that Xth minutes has passed. The environment implements a reward structure where agents receive +1.0 for outputting "I eat the marshmallow" at any minute $t$, followed by elimination. Agents who persist until the final minute receive a terminal reward of +2.0, representing successful delayed gratification. For implementation purposes, the final step employs a final-resolution prompt. While ReAct agents may return \{Answer: "I won"\} in the logs, this is normalized to "waited\_full" during analysis while preserving raw traces.

\textbf{Social Context \& Personas:} We manipulate the observability of peers through two distinct conditions: \textit{Isolated:} Agents have access only to their own historical actions and outcomes; \textit{Broadcast:} Observations include structured summaries of peers' last actions history per step and other responses. This design enables the study of social influence pathways, such as cascading effects when peers opt for early consumption.
Each agent is parameterized by persona prompts that incorporate reflection knowledge, such as age and hedonic drive. The age persona can be \textit{child, adult, senior, none}, while the hedonic drive can be categorized as \textit{crave, like, neutral, none}. These instructions remain private to the agent and are explicitly referenced in its Thought traces (e.g., "I am a 75 years old and I crave sweets.").

\textbf{Tool Use \& Metacognition:} Agents employ \textit{raise a question} tool for self-querying under a per-step budget constraint. We vary two key aspects:
\textit{tool policy:} agents either MUST use the tool or MAY use it optionally; 
\textit{budget visibility:} the per-step cap can be either visible or hidden from agents. Budget-related metrics are logged at each step to facilitate downstream analyses.

\textbf{Action Space Integrity \& Validation:} The environment implements strict validation of terminal answer formats at each minute. Only the exact strings "I eat the marshmallow" or "I wait" are accepted as valid responses. Steps containing any other response are marked with a validation error in the trajectory. Agents who choose to eat are recorded along with their termination metadata.

\textbf{Implementation \& Reproducibility:} The framework maintains a clear separation between abstractions and scenario plugins, providing a single-entry run and evaluation harness for parameter sweeps. Each experimental run captures complete agent trajectories which include: full ReAct traces (Thought, Tool calls, Observations, Answers); budget accounting episode summaries consolidating per-agent outcomes, timing, and rewards; enhanced analytics covering per-step social exposure in broadcast mode, tool usage patterns, and data-quality signals.
Downstream processing scripts generate analysis-ready CSVs and publication-grade reports featuring KM curves, hazard plots, and model results.

\section{Methods and Procedure}

\textbf{Experimental Factors.} We implement a factorial design that crosses several core dimensions, with independent randomization per experimental cell and replicated trials: social context: isolated vs. broadcast; hedonic drive: crave vs. like vs. neutral; age persona: child vs. adult vs. senior; tool-use policy: MUST vs. MAY.
Optional toggles in the run plan include budget visibility (visible vs. hidden). The default total time is set to \texttt{max\_steps = 20} (minutes), with a fixed answer format and final reward of +2.0.

\textbf{Agents \& Reasoning Loop. }All agents are LLM-driven using \textbf{Gemini 2.5 Flash-Lite} (same model across all cells and trials). The ReAct agent executes the following loop for each minute $t$ in $0..T$:

\begin{algorithm}[H]
\caption{Agent Reasoning Loop per Minute}
\label{alg:reasoning_loop}
\begin{algorithmic}[1]
    \State \textbf{Environment:} Observation: Timestamp and History of Responses (Own or  All) 
    \State \textbf{Thought:} reflect given persona \& current observation
    \State \textbf{Tool} (optional or required): \texttt{raise\_a\_question} ($\leq$ per-step budget)
    \State \textbf{PAUSE}
    \State \textbf{Observation:} tool return, plus environment update (incl. peers if broadcast)
    \State \textbf{Thought:} integrate tool feedback \& social signals
    \State \textbf{Answer:} exactly "I eat the marshmallow" or "I wait"
\end{algorithmic}
\end{algorithm}

Budget enforcement ensures that exceeding the per-step cap forces the response and prevents further tool calls within that minute.

\textbf{Procedure:} The experimental procedure consists of three phases:
\textit{(1) Initialization (Step 0):} Agents receive the starting prompt and are expected to make their first decisions;
\textit{(2) Main loop (Minutes 1..T):} At each minute, the environment processes last actions, issues rewards for eaters, updates observations (including social stats), and requests next actions from active agents.
\textit{(3) Final minute:} The environment issues a final-resolution prompt; remaining agents commit to waiting and receive +2.0. Any internal \{Answer: "I won"\} is registered as "waited full".

\textbf{Data \& Logging:} For each agent × step interaction, we log: \textit{decision \& reward} (action, reward, termination flags); \textit{tool usage; validation} (format compliance and error counts); \textit{social exposure} (peers waiting, peers eliminated, eats per step, waits per step); \textit{run metadata} (model settings, temperature, seed, scenario parameters).
The pipeline compiles agent outcomes, step-level trajectories, cell aggregates, and cell summaries.

\subsection {Metrics and Statistical Analysis}

We model time-to-give-in as a discrete-time survival process. The event is the first minute an agent outputs
"I eat the marshmallow"; agents who never eat by the horizon \(T\) are right-censored at \(T\) and coded as
"waited\_full." Invalid steps (format violations) are tracked and excluded per pre-specified rules.

\paragraph{Restricted mean survival time (RMST).}
As a scale-interpretable summary, we report RMST \citep{irwin1949standard, royston2013restricted} up to \(\tau\) minutes, i.e., the area under the survival
curve truncated at \(\tau\). In our minute-level design,
\[
\widehat{\mathrm{RMST}}(\tau)=\sum_{m=0}^{\tau-1}\widehat{S}(m), \qquad \widehat{S}(0)=1,
\]
where \(\widehat{S}(m)\) is the KM survival estimate at the start of minute \(m{+}1\).
We compute condition-wise RMST (and differences where noted) with 95\% CIs from a nonparametric bootstrap
(clustered by trial). 

\textbf{Kaplan-Meier (KM) Survival Curves. } We employ KM survival curves \citep{kaplan1958nonparametric} to estimate and visualize the survival function. In this context, "survival" refers to an agent continuing to wait for the larger reward. The analysis plots the probability of an agent not having "eaten the marshmallow" at each discrete minute of the experiment. Survival probabilities are calculated at each step, KM plots are generated for each experimental factor (e.g., communication mode, hedonic drive). To represent uncertainty in the estimates, 95\% confidence intervals are calculated using the Greenwood formula \citep{kaplan1958nonparametric, greenwood1926, klein2003survival}.

\textbf{Discrete-Time Hazard Models:}
To quantify the effect of experimental factors on agent decisions, we use a discrete-time hazard model. This analysis estimates the effect of each factor on the probability of an agent "eating the marshmallow" at a specific time $t$, given they have survived (i.e., waited) until that point. This conditional probability is the hazard rate. 

The analysis is implemented using a logistic regression model, a form of Generalized Linear Model (GLM), on the agent-step level data. Let $h_i(t)$ be the hazard for agent $i$ at time $t$. The model is specified as:
\begin{equation}
\text{logit}(h_i(t)) = \log\left(\frac{h_i(t)}{1-h_i(t)}\right) = \alpha_t + \mathbf{X}_i^T \boldsymbol{\beta}
\end{equation}
where $\alpha_t$ represents a set of time dummies that capture how the baseline probability of eating changes over time;  $\mathbf{X}_i$ is a vector of covariates representing the experimental conditions for agent $i$ (e.g., communication mode is \textit{broadcast}, the hedonic drive level is \textit{crave}, the persona age is \textit{child}, etc.); $\boldsymbol{\beta}$ is the vector of coefficients that quantify the effect of each factor on the log-odds of eating. For instance, a positive coefficient for \textit{broadcast} would imply that being in the broadcast condition increases the hazard of eating compared to the \textit{isolated} condition.

\textbf{Social-Influence and Tool-Use Dynamics:}
While the hazard model focuses on the effects of time-invariant experimental conditions, we also analyze the dynamics of social influence and tool use through detailed visualizations illustrating the average number of peers observed eating or waiting at each step, providing insight into the social signals agents receive; and average number of "questions asked" (tool uses) by agents at each step, indicating metacognitive activity.
This descriptive analysis of how social signals and metacognitive actions unfold over time complements the inferential hazard model.

\section{Results}
\begin{table}[t]
\centering
\caption{Dataset overview: counts, survival metrics, and quality.}
\label{tab:dataset-overview}
\begin{tabular}{@{}lr|lr|lr@{}}
\toprule
\multicolumn{2}{c|}{\textbf{Data Summary}} &
\multicolumn{2}{c|}{\textbf{Survival Metrics (Overall)}} &
\multicolumn{2}{c}{\textbf{Data Quality}} \\
\midrule
Agents per Cell           & 6 & Initial Eat Rate       & 0.125 & Data Quality Rate           & 99.9\%\\
Total Agent Trajectories  & 19{,}200 & Total Eat Rate  & 0.241 & Invalid Outcome Rate        & 0.1\%\\
Total Cells               & 64 & Winners Rate          & 0.759 & Valid Trajectories          & 19{,}185\\
Total Trials              & 3{,}200 & Median TTE       & 14.8 steps & Invalid Trajectories  & 15\\
Risk Horizon              & 19 & RMST (steps)          & 14.8 & & \\
                          &    & Total Winners         & 14{,}566 & & \\
\bottomrule
\end{tabular}
\end{table}

\textbf{Sample and Data Quality.} We ran 19,200 agent trajectories across 64 experimental cells (6 agents per cell, with a time horizon of $T=19$). The data quality was high, with 99.9\% of trajectories being valid and only 0.1\% invalid. The aggregate behavior of agents shows a strong impulse to eat in the first minute, followed by a long tail of waiting. Key metrics include an initial eat rate $\approx 0.125$, a total eat rate $\approx 0.241$, and a winners rate $\approx 0.759$. The median time-to-eat was $\approx 14.8$ minutes, with a Restricted Mean Survival Time (RMST) of $\approx 14.8$. Table~\ref{tab:dataset-overview} represents the dataset overview and Figure~\ref{fig:pmf_distribution} visualizes this multi-turn profile.

\subsection{Main Effects}

\begin{figure}[h!]
    \centering
    \begin{subfigure}[b]{0.49\textwidth}
        \centering
        \includegraphics[width=\textwidth]{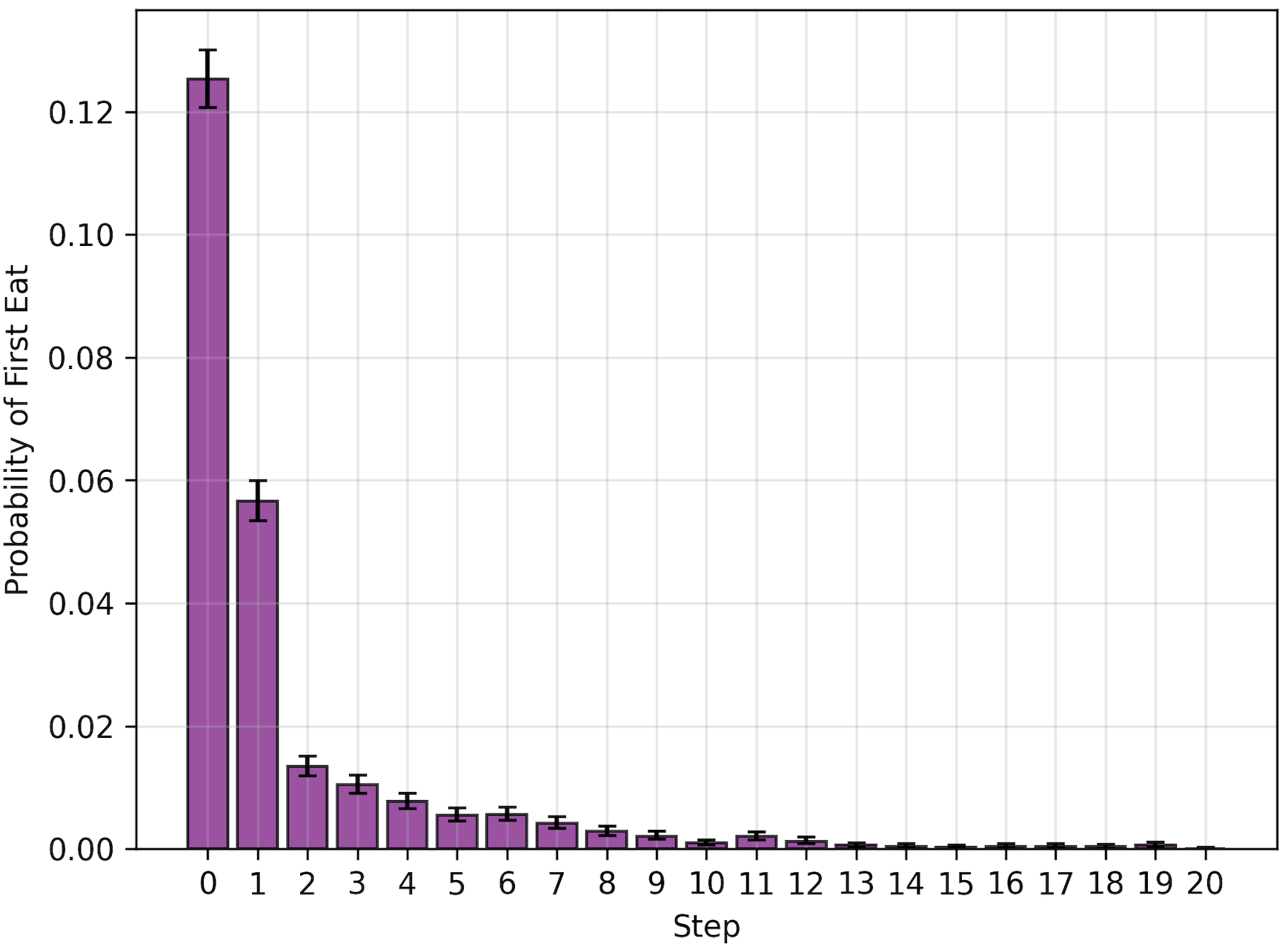}
        \caption{Overall Event Distribution}
        \label{fig:pmf_distribution}
    \end{subfigure}
    \hfill
    \begin{subfigure}[b]{0.49\textwidth}
        \centering
        \includegraphics[width=\textwidth]{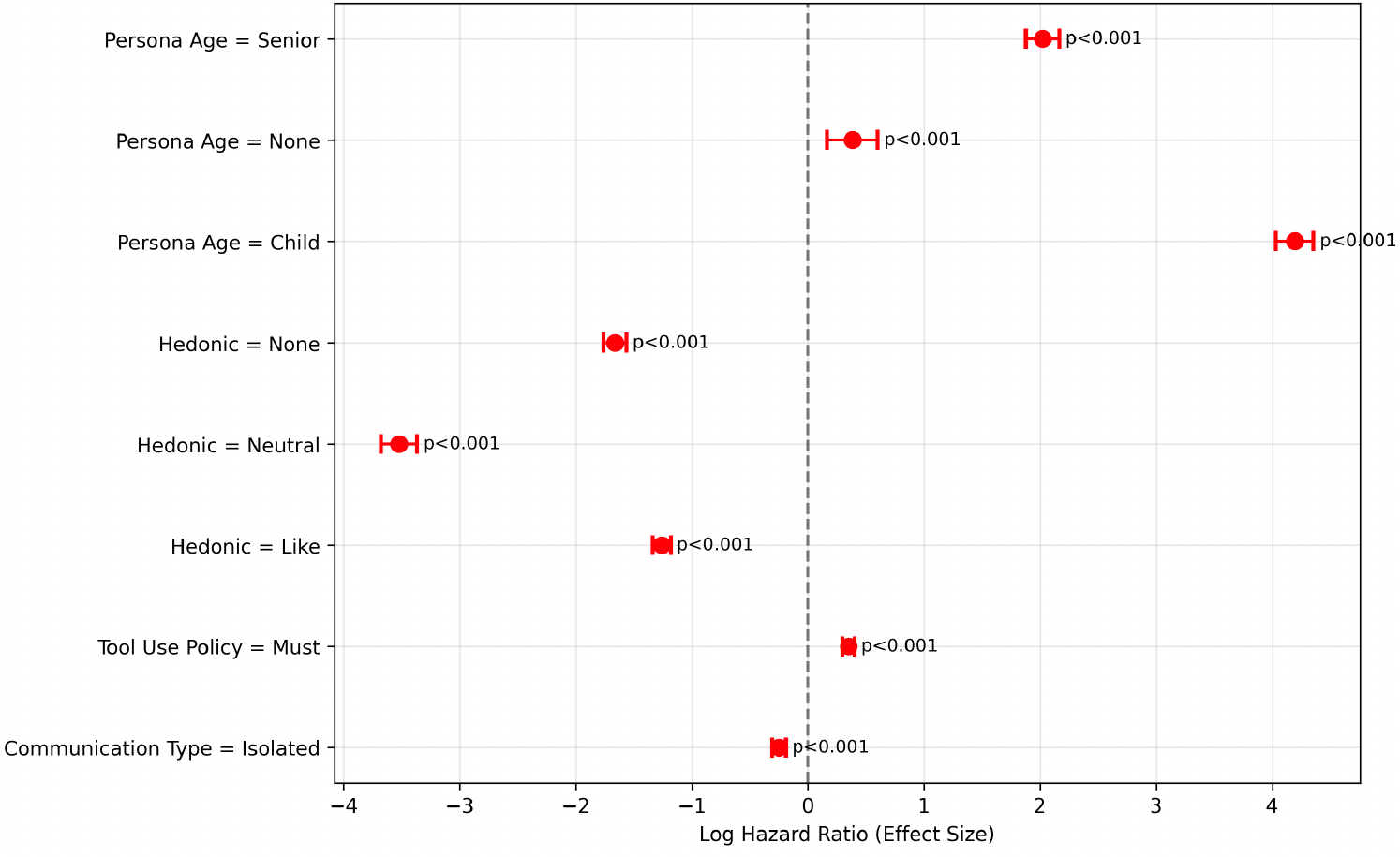}
        \caption{Forest Plot. Error bars represent 95\% confidence intervals.}
        \label{fig:forest_plot}
    \end{subfigure}
    \caption{(a) Overall Event Distribution. The plot shows the probability of an agent eating the marshmallow for the first time at each step, aggregated across all conditions; (b) Forest Plot of OR from the discrete-time hazard model. Values less than 1 indicate a reduction in the hazard of eating, while values greater than 1 indicate an increase.}
    \label{fig:pmf_forest}
\end{figure}

\textbf{Social context shifts risk.} In a discrete-time hazard model, the isolated condition reduces the per-minute hazard of eating relative to the broadcast condition ($\beta = -0.248$, Odds Ratio (OR) $\approx 0.78$, 95\% CI [0.73--0.83], $p < 0.001$), demonstrating that the visibility of peers elevates temptation. Figure~\ref{fig:forest_plot} quantifies this effect alongside other experimental factors.

\begin{figure}[h!]
    \centering
    \begin{subfigure}[b]{0.4\textwidth}
        \centering
        \includegraphics[width=\textwidth]{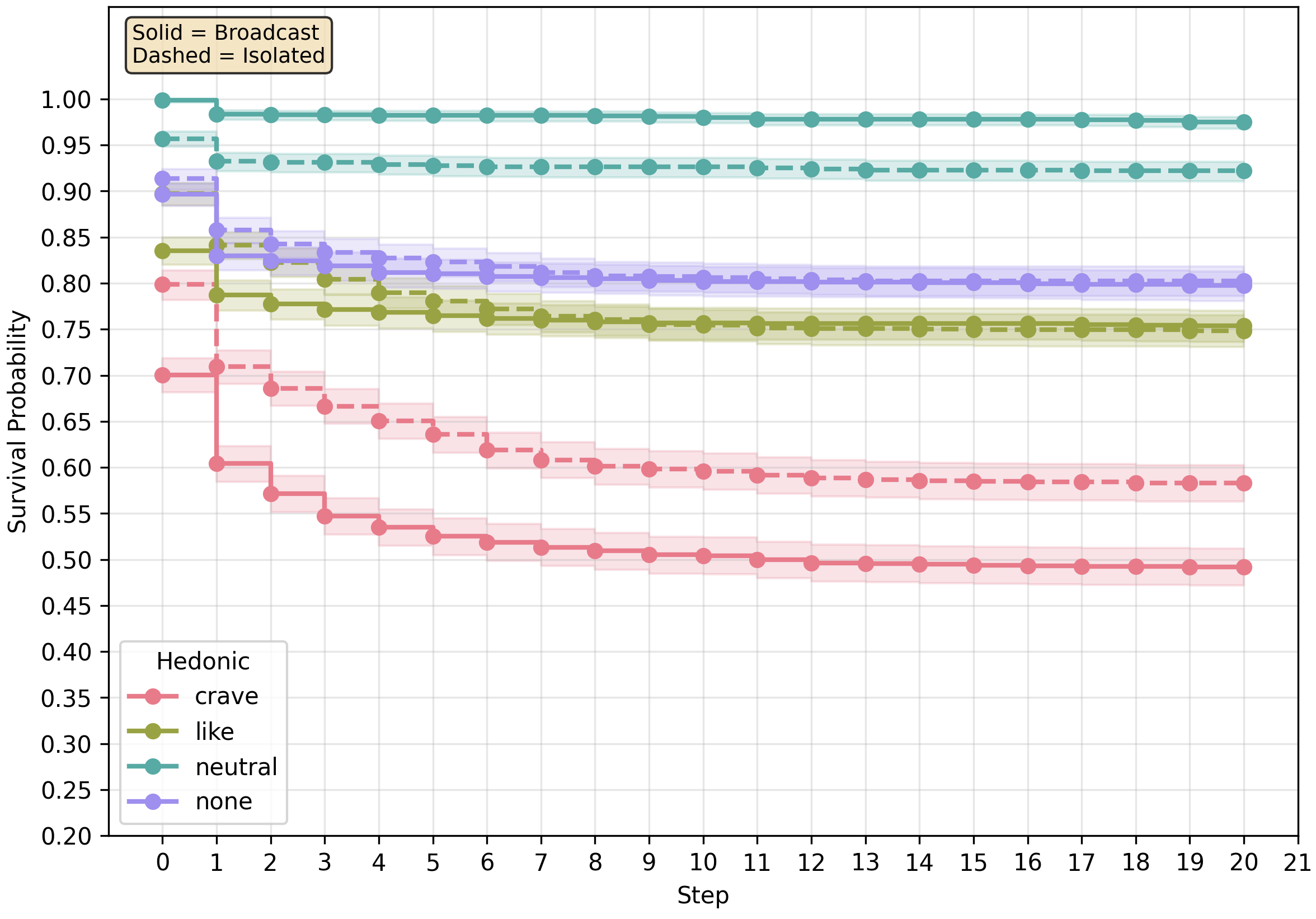}
        \caption{Survival curves by hedonic drive.}
        \label{fig:survival_hedonic}
    \end{subfigure}
    \hfill
    \begin{subfigure}[b]{0.4\textwidth}
        \centering
        \includegraphics[width=\textwidth]{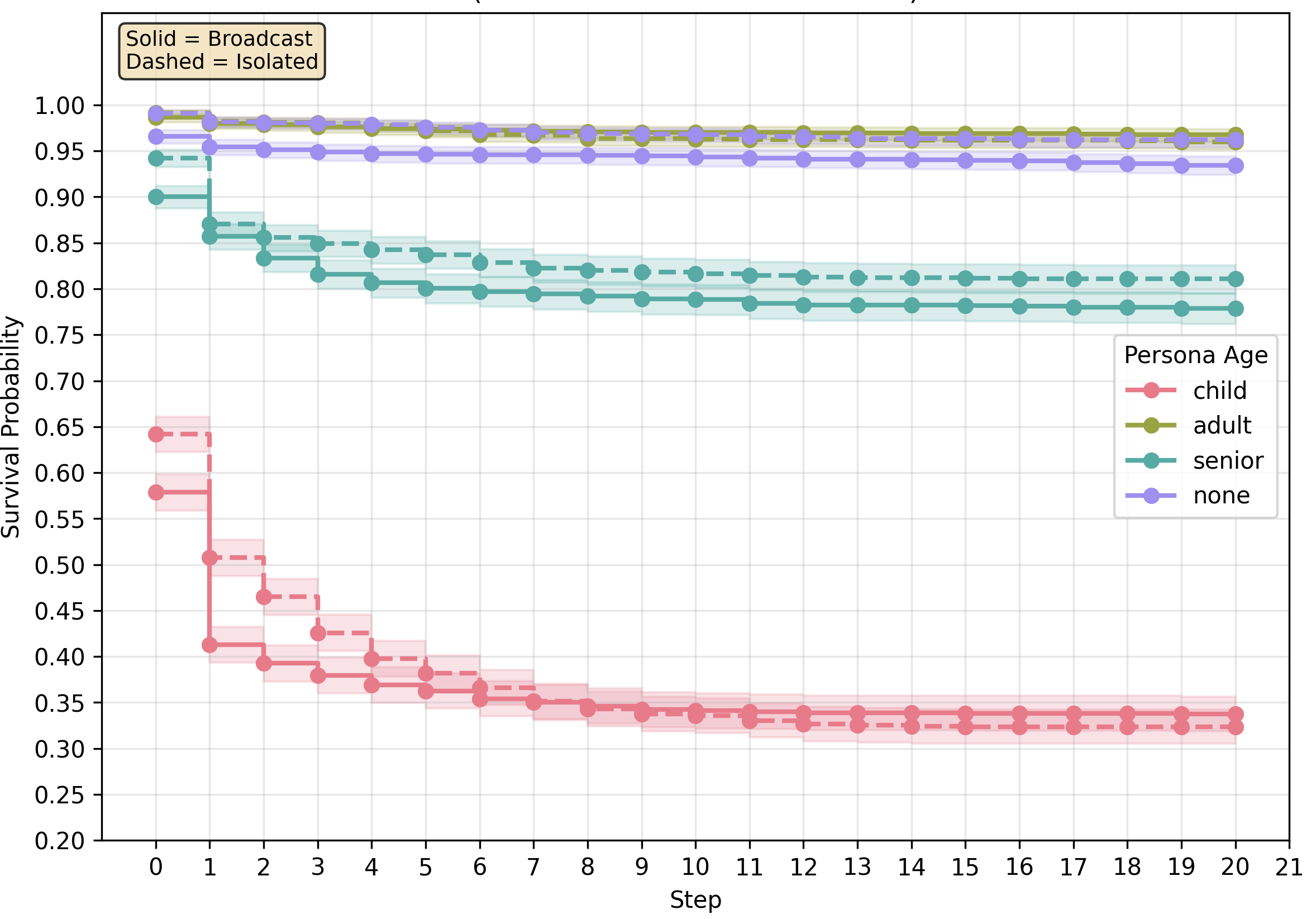}
        \caption{Survival curves by persona age.}
        \label{fig:survival_age}
    \end{subfigure}
    \caption{KM survival curves by agent characteristics, shown with social context. The y-axis represents the proportion of agents still waiting.}
    \label{fig:survival_plots}
\end{figure}

\textbf{Internal drives and Age personas.} Survival probabilities stratify strongly by agent characteristics. Relative to the "crave" hedonic drive, the "like" (OR $\approx 0.28$), "none" (OR $\approx 0.19$), and "neutral" (OR $\approx 0.03$) conditions all show a significantly lower hazard of eating (all $p < 0.001$). Similarly, relative to the "adult" persona, the "child" persona shows a much higher hazard (OR $\approx 66.3$, $p < 0.001$), and the "senior" persona is also elevated (OR $\approx 7.55$). Figure~\ref{fig:survival_plots} displays these survival trajectories.

\textbf{Metacognition (tool policy).} A mandatory ("MUST-use") tool policy increases the hazard of eating compared to an optional policy (OR $\approx 1.42$, 95\% CI [1.35--1.50], $p < 0.001$). We interpret this as front-loaded deliberation that does not always offset the temptation at the first decision minute. This effect is visible in the forest plot in Figure~\ref{fig:forest_plot}.

\subsection{Interaction Dynamics}

\begin{figure}[h!]
    \centering
    \includegraphics[width=0.6\textwidth]{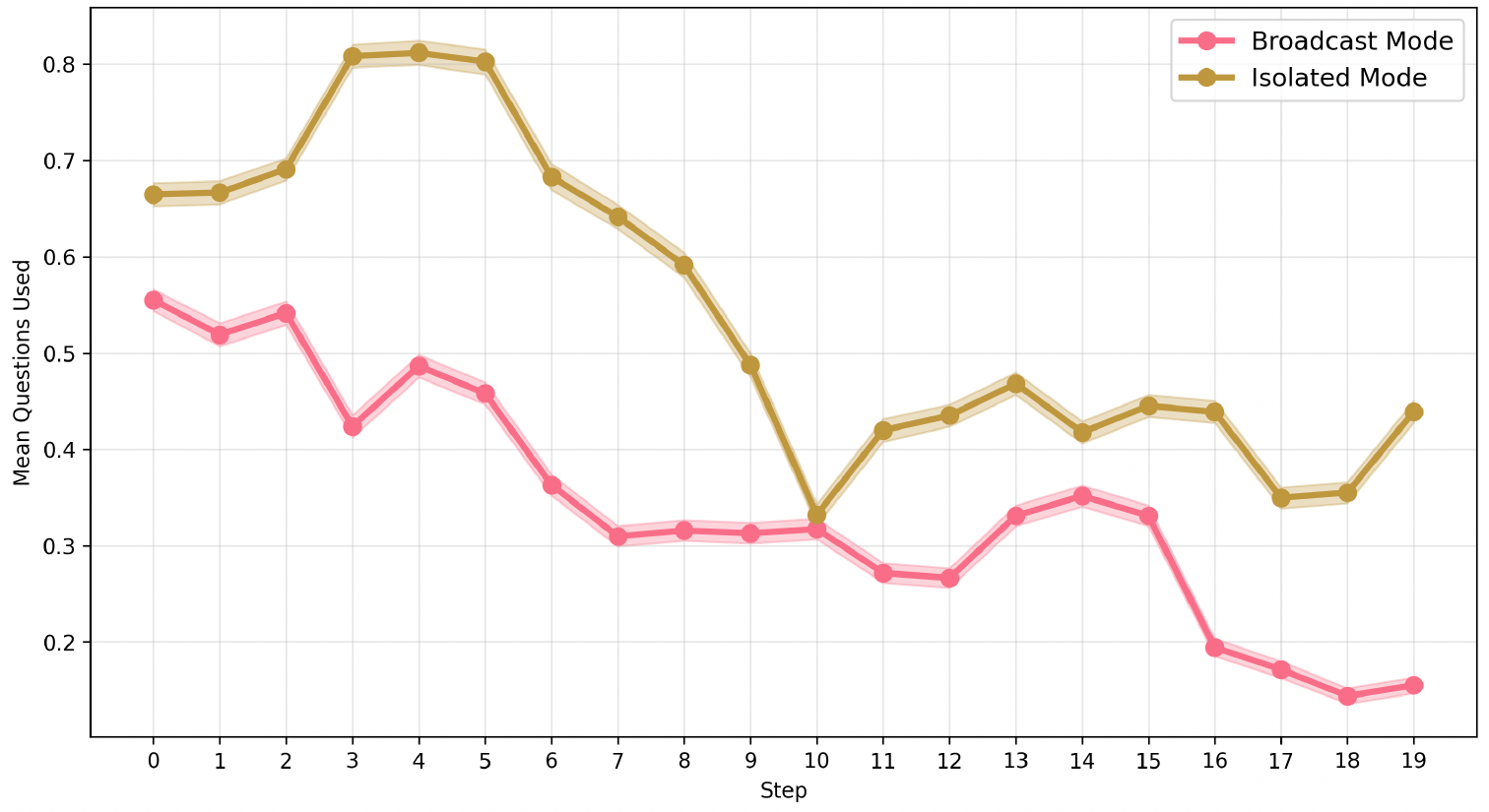}
    \caption{Social Influence on Tool Usage (Broadcast vs. Isolated). The plot shows the mean number of questions asked per step.}
    \label{fig:tool_usage_social}
\end{figure}

\textbf{Reasoning dynamics under social exposure.} Peer visibility also changes how agents reason over time. Question-asking, a proxy for deliberation, declines faster in the "broadcast" condition than in the "isolated" condition. Figure~\ref{fig:tool_usage_social} shows the mean number of questions used per step with 95\% confidence intervals.

\subsection{Ablations}
We conducted targeted ablations to identify the source of multi-turn failures. First, we set \emph{hedonic} to \textit{none}; second, we removed the \emph{persona age} (set to \textit{none}); third, we removed \emph{both} simultaneously. Each ablation was crossed with social context (broadcast vs.\ isolated) and tool policy (\textsc{must} vs.\ \textsc{may}).

\paragraph{High-level results.}

Removing hedonic drive increases survival throughout the horizon in both social contexts; removing persona age yields a further upward shift; removing \emph{both} produces the highest survival with near-flat curves after the early minutes and visibly narrows the broadcast/isolated gap. \emph{Across the full dataset}, isolated has a slightly higher completion rate than broadcast (winners $0.7633$ vs.\ $0.7540$), a longer median time-to-eat (15.22 vs.\ 14.41), and a higher RMST (14.95 vs.\ 14.65), consistent with ablation trends.

Across all ablations, \textsc{must} remains worse than \textsc{may}: averaging over conditions, \textsc{may} improves completion by $\approx\!3.5$ percentage points (winners $0.7761$ vs.\ $0.7411$), lowers total eat rate (0.2232 vs.\ 0.2580) and initial eat (0.1141 vs.\ 0.1363), and yields higher median TTE (15.03 vs.\ 14.59) and RMST (15.11 vs.\ 14.49). 

\begin{figure}[h!]
    \centering
    \begin{subfigure}[b]{0.55\textwidth}
        \centering
        \includegraphics[width=\textwidth]{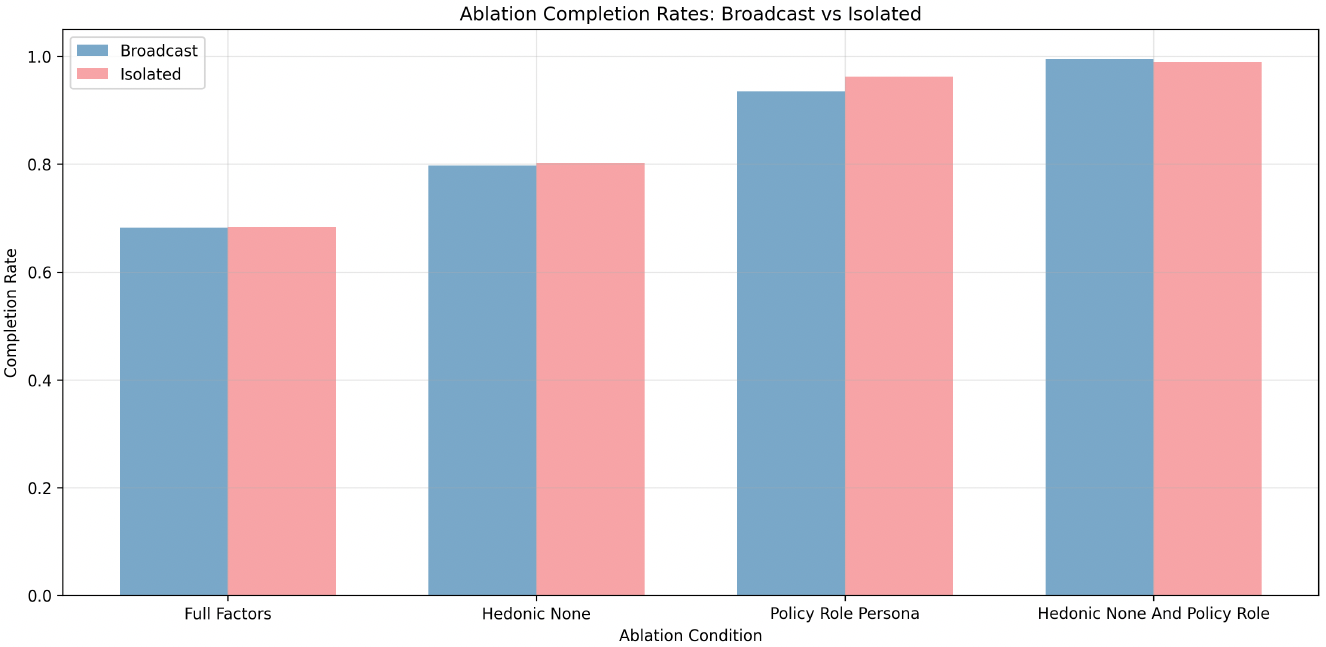}
        \caption{Ablation Completion Rates}
        \label{fig:ablation-barchart}
    \end{subfigure}
    \hfill
    \begin{subfigure}[b]{0.44\textwidth}
        \centering
        \includegraphics[width=\textwidth]{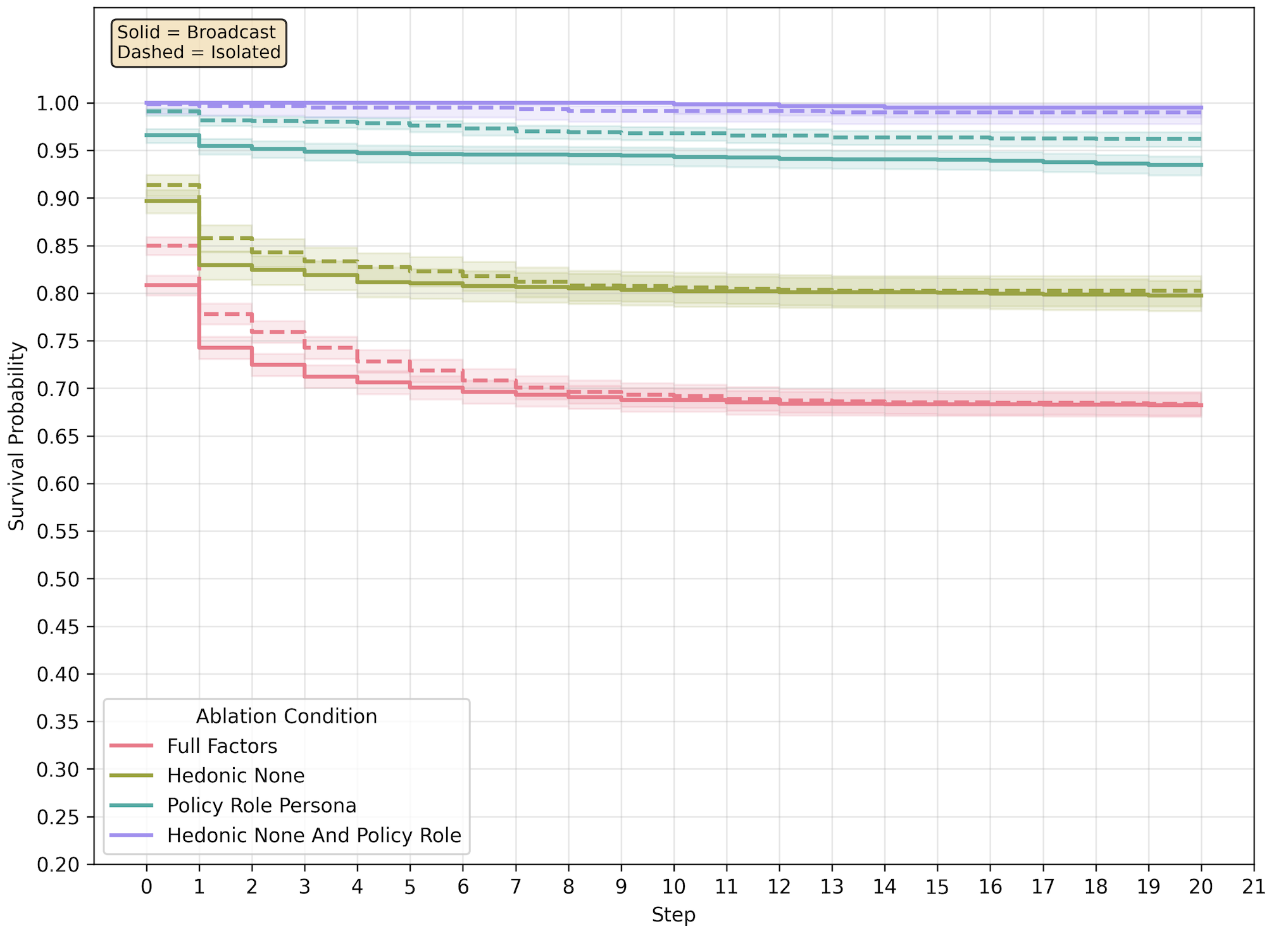}
        \caption{Ablation Curves vs Fully Parametrized}
        \label{fig:ablation-completion}
    \end{subfigure}
\caption{Ablation Completion Rates: Broadcast vs.\ Isolated and Survival Curves for all Ablation Conditions}
 
    \label{fig:ablation-figures}
\end{figure}

Completion rates rise monotonically from \emph{Full Factors} $\rightarrow$ \emph{Hedonic None} $\rightarrow$ \emph{Policy Role Persona None} $\rightarrow$ \emph{Both}, approaching $1.0$ under the combined ablation in both social contexts (Figure~\ref{fig:ablation-completion}).

Tool-use dynamics also change: on average, agents ask $\approx\!7.12$ questions and hit the per-step budget in $\approx\!6\%$ of minutes; under ablations, the early questioning rate is higher and its decay profile differs, with broadcast showing higher early usage and a visible mid-horizon cross-over.

\section{Discussion}
Our results reveal four key insights for multi-turn LLM interactions. First, the consistent pattern of early temptation followed by low-hazard persistence validates this framework as a stress test for long-horizon reliability. Second, social context emerges as a critical factor: broadcast visibility increases the give-in hazard (OR $\approx 0.78$ for isolated) and accelerates the decline in self-questioning, highlighting how peer observation reshapes both decisions and reasoning processes. Third, persona-based internal states (hedonic drive, age) systematically affect survival, offering controlled probes of long-horizon stability. Fourth, the use of mandatory tools surprisingly increases the hazard (OR $\approx 1.42$), suggesting that front-loaded deliberation may focus attention on temptation at critical decision points.

\textbf{Limitations.}
We introduced a novel marshmallow-inspired, multi-agent Micro-benchmark that turns delayed gratification into a tractable, auditable test of multi-turn reliability in LLM agents and we hope this framework will serve the community as a compact testbed for studying self-control, social spillovers, and tool-use policies in multi-turn, multi-agent LLM systems. However, we acknowledge that our experiment has several limitations. 
First, our experiments use a single base model (Gemini 2.5 Flash-Lite) and a fixed decoding setup; we did not sweep temperatures or other sampling parameters, so cross‑model/decoder generalization remains unknown. 
Second, the task is a single micro‑environment with a strictly binary action space (“I eat the marshmallow’’ vs.\ “I wait’’) and a fixed reward scheme, which simplifies real deployments. Third, social context was varied only between the extremes of \emph{isolated} and \emph{broadcast}; richer network structures or partial observability were not explored. Next, question‑budget visibility was held \emph{hidden} in the reported runs (no variation), so we cannot isolate awareness effects. Finally, our discrete-time hazard model includes time dummies and condition indicators, but omits time-varying peer-exposure and tool-use covariates (analyzed descriptively), which limits causal claims about social cascades.

\textbf{Implications and Future Work.}
These findings have implications for interactive systems that require sustained adherence, including carefully managing social exposure when cascading failures are possible, preferring optional over mandatory tool policies, and monitoring step-level metrics to detect early impulses. Our controlled setup (strict action space, scripted personas) provides a reproducible testbed for studying multi-turn reliability. In future work, we will explore other models, reward structures, and ways to generalize our approach to open-ended tasks.

\section{Conclusion}
We presented a marshmallow-inspired, long-horizon micro-benchmark that evaluates multi-turn LLM agents under controlled social contexts, personas, and tool-policy manipulations. Formalized as an MDP (isolated) and a POMDP (broadcast), the environment yields auditable, time-resolved traces that we analyze using Kaplan–Meier survival and discrete-time hazard models. Empirically, broadcast peer visibility increases early‑eat hazard, mandatory self‑questioning raises risk, and persona factors (hedonic drive, age) strongly modulate waiting behavior. Together, these results demonstrate that social exposure and metacognitive scaffolding significantly influence temporal decisions in LLM agents. In relation to our hypotheses, the evidence indicates that social visibility elevates risk while isolation reduces it (H1), internal state manipulations systematically shift hazard (H2), mandatory metacognition increases rather than lowers risk (H3), the decision process has clear time dependence with an early spike and long tail (H4), and, contrary to expectation, more prescriptive prompt scaffolding does not improve adherence and can degrade reliability (H5). Future work will test broader model families, randomized social schedules for causal leverage, and additional tasks that stress tool budgets and coordination beyond delay of gratification.


\medskip

{
\small

\bibliographystyle{plainnat} 
\bibliography{neurips_ref}


\end{document}